\documentclass{article}

\usepackage{ijcai25}

\usepackage{times}
\usepackage{soul}
\usepackage{url}
\usepackage[hidelinks]{hyperref}
\usepackage[utf8]{inputenc}
\usepackage[small]{caption}
\usepackage{graphicx}
\usepackage{amsmath}
\usepackage{amsthm}
\usepackage{booktabs}
\usepackage{algorithm}
\usepackage{algorithmic}
\usepackage[switch]{lineno}

\usepackage{natbib}
\usepackage{caption}
\usepackage{enumitem}
\usepackage{amsfonts}
\usepackage{amsmath}
\usepackage{amsthm}
\usepackage[table,xcdraw]{xcolor}

\theoremstyle{definition}

\usepackage{dsfont} 
\usepackage{multirow} 
\usepackage{amssymb}

\usepackage{newfloat}
\usepackage{listings}
\usepackage{longtable}
\usepackage{booktabs} 

\title{SetKE: Knowledge Editing for Knowledge Elements Overlap}

\author{
 \textbf{Yifan Wei\textsuperscript{1,2}},
 \textbf{Xiaoyan Yu\textsuperscript{1,3}},
 \textbf{Tengfei Pan\textsuperscript{2}},
 \textbf{Angsheng Li\textsuperscript{1}\footnotemark[2]} , 
 \textbf{Li Du\textsuperscript{2}\footnotemark[2]}
 \\
\textsuperscript{1}State Key Laboratory of CCSE,School of Computer Science and Engineering,Beihang University\\
\textsuperscript{2}Beijing Academy of Artificial Intelligence,
\textsuperscript{3}Beijing Institute of Technology
 \\
 \texttt{weiyifan@buaa.edu.cn, 
 \{tfpan,duli\}@baai.ac.cn}
 }
 
\author{
Yifan Wei$^1$
\and
Xiaoyan Yu$^2$\footnotemark[2]\and
Ran Song$^{3}$\and
Hao Peng$^{1}$\And
Angsheng Li$^1$\footnotemark[2]\\
\affiliations
$^1$State Key Laboratory of CCSE, School of Computer Science and Engineering, Beihang University\\
$^2$Beijing Institute of Technology, 
$^3$Kunming University of Science and Technology
\emails
\{weiyifan, angsheng\}@buaa.edu.cn,
xiaoyan.yu@bit.edu.cn,
song\_ransr@163.com
}

\begin{document}

\maketitle

\begin{abstract}
Large Language Models (LLMs) excel in tasks such as retrieval and question answering but require updates to incorporate new knowledge and reduce inaccuracies and hallucinations. Traditional updating methods, like fine-tuning and incremental learning, face challenges such as overfitting and high computational costs. Knowledge Editing (KE) provides a promising alternative but often overlooks the Knowledge Element Overlap (KEO) phenomenon, where multiple triplets share common elements, leading to editing conflicts.
We identify the prevalence of KEO in existing KE datasets and show its significant impact on current KE methods, causing performance degradation in handling such triplets. To address this, we propose a new formulation, Knowledge Set Editing (KSE), and introduce SetKE, a method that edits sets of triplets simultaneously. Experimental results demonstrate that SetKE outperforms existing methods in KEO scenarios on mainstream LLMs. Additionally, we introduce \textsc{EditSet}, a dataset containing KEO triplets, providing a comprehensive benchmark.
\end{abstract}

\renewcommand{\thefootnote}{\fnsymbol{footnote}}
\footnotetext[2]{Corresponding Authors.}

\section{Introduction}

    Large Language Models (LLMs) function as powerful knowledge repositories, excelling in tasks like retrieval and question answering \citep{petroni2019language,geva2020transformer}.
    However, the dynamic nature of factual information necessitates ongoing updates to prevent inaccuracies and hallucinations post-deployment.
    While parameter-efficient fine-tuning and incremental learning techniques offer ways to update knowledge in LLMs, these paradigms can lead to potential drawbacks such as overfitting and significant computational costs.
    In response to these challenges, the concept of Knowledge Editing (KE), which involves directly modifying specific pieces of knowledge within the LLMs without extensive retraining, has emerged as a focal point \citep{dong2022calibrating,wei2023assessing,gupta2023editing,song2024does,yao2024knowledge}.

\begin{figure}[t]
    \centering
    \includegraphics[width=1\linewidth]{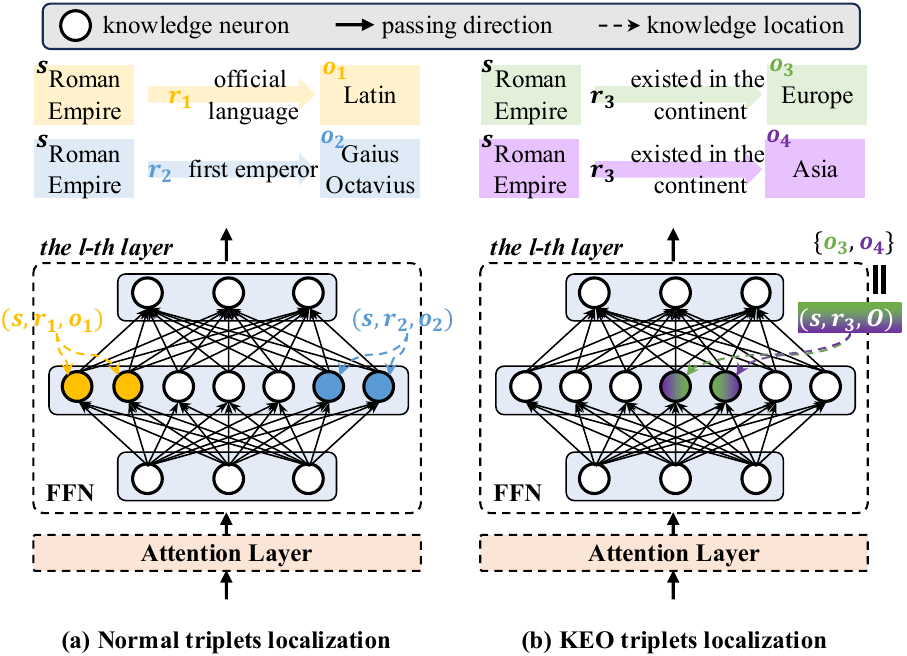}
    \caption{\label{fig:showcase}
    A demonstration of normal triplets and KEO triplets, along with a toy example showing how they are localized within Transformer-based LLMs, where normal triplets are mapped to distinct neurons, and KEO triplets are mapped to overlapping neurons.}
\end{figure}

    Current research on KE formalizes factual knowledge within LLMs as triplets $(s, r, o)$, consisting of a subject $s$, an object $o$, and their relation $r$. 
    These studies can be classified into three main approaches: \textbf{1) Vanilla Editing} focuses on the ability to modify a single factual knowledge entry \citep{sinitsin2019editable,de2021editing,rawat2021modifying,dai2022knowledge}. 
    They make a single modification to model parameters with no subsequent changes.
    \textbf{2) Sequential Editing} involves iteratively updating knowledge triplets one at a time \citep{hartvigsen2024aging,hu2024wilke,yu2024melo,wang2024wise}. 
    They support streaming edits, ensuring that knowledge updates are continuous and orderly.
    \textbf{3) Batch Editing} enables simultaneous editing of multiple knowledge instances, with each instance assigned a unique editing objective (e.g., $o \rightarrow o^*$) \citep{mitchell2022memory,tan2023massive,li2024pmet}.
    Overall, current methods focus on situations where a knowledge prefix $t_r(s)$ (comprising the subject $s$ and the relation $r$) corresponds to a single object, treating it as a singular target for editing. 
    As illustrated in Figure \ref{fig:showcase}(a), the two triplets have distinct relations and objects, resulting in different prefixes $t_{r_1}(s)$ and $t_{r_2}(s)$, which current approaches can effectively handle.

    In real-world scenarios, a single subject often corresponds to multiple objects under the same relation.
    For instance, as illustrated in Figure \ref{fig:showcase}(b), the Roman Empire spanned multiple continents, resulting in triplets that share the same subject and relation but differ in their objects. 
    We refer to these as knowledge element overlap (KEO) triplets.
    In this context, the editing knowledge prefix $t_{r_3}(s) = t_{r_4}(s)$, such as "What continent did the Roman Empire exist in?", corresponds to multiple valid answers, including 
    \textit{Europe} and \textit{Asia}.
    Current methods often overlook this kind of overlap, which is problematic as knowledge triplets with shared elements within the same factual context require simultaneous consideration to ensure consistency during knowledge editing.

    To further investigate the consequences of this overlook, we analyze existing mainstream KE datasets to evaluate the prevalence of Knowledge Element Overlap (KEO), which we found widespread.
    We then separate instances with KEO from those without and conduct experiments to evaluate the performance of current KE methods on these instances. 
    As anticipated, the results reveal a significant performance decline for instances with KEO. 
    Further analysis reveals that triplets with overlapping elements share knowledge neurons in the FFN layers, as shown in Figure \ref{fig:showcase}(b), leading editors to overwrite knowledge unintentionally. 
    For example, modifying \textit{Europe} to \textit{Africa} may inadvertently cause both \textit{Europe} and \textit{Asia} to be altered to \textit{Africa}.
    Building on this observation, we collect KEO instances from Wikidata to construct a new dataset, \textsc{EditSet}, enabling a more comprehensive exploration of KEO in KE. 
    The dataset comprises over 700 relation types, with our study focusing on the 31 most common ones, consistent with prior research \citep{levy2017zero,elazar2021measuring,meng2022locating,zhong2023mquake,wei2024does,yin2024history,maneighboring}.
    
    Given that the current methods have failed to effectively handle the KEO situation, this paper
    introduce an editing framework named \underline{\textbf{Set}} \underline{\textbf{K}}nowledge \underline{\textbf{E}}ditor (\underline{\textbf{SetKE}}), which employs bipartite matching for optimization to enable knowledge editing in scenarios involving overlapping knowledge elements. 
    Experimental results demonstrate that SetKE significantly outperforms existing KE methods in KEO scenarios. 
    Furthermore, in-depth analyses confirm that SetKE effectively mitigates knowledge overwriting. 
    Our main contributions are summarized as follows:
    
    $\bullet$ We propose a novel formulation of Knowledge Set Editing (KSE) and construct a new dataset, \textsc{EditSet}, to facilitate in-depth exploration of Knowledge Element Overlap (KEO).
    
    $\bullet$ We introduce a new set editing framework, Set Knowledge Editor (SetKE), leveraging bipartite matching to achieve the editing target by treating knowledge entries as a set.
    
    $\bullet$ Extensive experiments show that SetKE significantly outperforms existing methods in KEO scenarios, establishing it as a powerful paradigm for editing knowledge sets.

\section{Preliminaries}

    This section outlines the definition of Knowledge Element Overlap (KEO), the current task formulation for Knowledge Editing, and our formulation for Knowledge Set Editing.

    \subsection{Knowledge Element Overlap}

        A factual knowledge entry $\mathcal{K}$ can be represented as a triplet $(s,r,o)$, where the knowledge elements refer to the subject $s$, the object $o$, and their relation $r$. 
        Based on the degree of overlap among knowledge elements—defined as sharing the same $s$, $o$ or $r$—knowledge triplets can be classified into four types (inspired by the categorization by \citet{sui2023joint}):
        \begin{itemize}[leftmargin=15pt]
            \item Normal: At most one knowledge element overlaps.
            \item Subject Object Overlap (\textit{SOO}): Overlapping subject-object pair ($s$ and $o$) is shared among some triplets.
            \item Relation Subject Overlap (\textit{RSO}): Overlapping subjects and shared relations among some triplets.
            \item Relation Object Overlap (\textit{ROO}): Overlapping objects and shared relations among some triplets.
        \end{itemize}
        Apart from the Normal type, all other types are considered instances of knowledge element overlap (KEO).

    \subsection{Knowledge Editing}
    
        The widely adopted formulation of Knowledge Editing \citep{geva2021transformer,geva2022transformer} represents knowledge $\mathcal{K}$ is stored in the language model $f_{\theta}$ in the form of triplets $(s, r, o)$. 
        The objective of KE is to modify the model $f_{\theta}$ into $f_{\theta}^*$ to achieve the transformation $(s, r, o)\rightarrow (s,r,o^*)$, whereby an input $x = t_r(s)$ is associated with its post-edit output $y = o^*$.
        Here, $t_r(s)$, referred to as the knowledge prefix, is a template used to describe the relation $r$ with the subject $s$.

        However, KEO is prevalent in real-world scenarios, posing challenges to the current KE framework, which uses a knowledge prefix $t_r(s)$ to edit a single object. 
        In this context, the RSO type of KEO impacts the editing result, as other types generate different prefixes $t_r(s)$.
        Therefore, in this paper, KEO typically refers to the RSO type.
        In this case, multiple triplets share the same prefix $t_r(s)$, represented as $O=\{o_1, o_2, \dots, o_N\}$, where $N$ denotes the number of distinct objects.
        The current KE formulation fails to distinguish between these scenarios: $(s, r, o)\rightarrow (s,r,o^*)$ and $(s, r, O)\rightarrow (s,r,O^*)$.
        In the latter case, modifying a single object is often insufficient to achieve the desired outcome.
        Consequently, existing KE methods perform poorly, as proven in Section \ref{sec:pilot}. 
        This highlights the need for a new formulation to address these limitations.

    \subsection{Our Formulation: Knowledge Set Editing}
        
        In our new formulation of Knowledge Set Editing (KSE), the editing target is defined as 
        $(s,r,O)\rightarrow (s,r,O^*)$, where the object is a set of entities ${O^*}=\{{o_1^*},{o_2^*},\ldots,{o_N^*}\}$.   
        In the KEO scenarios, such as $\langle \text{\textit{Roman Empire}}, \text{\textit{continent}}, \text{\textit{Europe}} \rangle$ and $\langle \text{\textit{Roman Empire}}, \text{\textit{continent}}, \text{\textit{Asia}} \rangle$, indicating the Roman Empire spanned multiple continents, editing such knowledge (the transformation $O \rightarrow O^*$) requires set operations. 
        For instance, consider the current values of $s=\text{\textit{Roman Empire}}$, $r=\text{\textit{existed in the continent}}$, and $O=\{\text{\textit{Europe}},\text{\textit{Asia}}\}$. 
        To modify $\text{\textit{Europe}}$ to $\text{\textit{Africa}}$, the more reliable way is to perform $(O=\{\text{\textit{Europe}},\text{\textit{Asia}}\})\rightarrow (O^*=\{\text{\textit{Africa}},\text{\textit{Asia}}\})$ instead of $(o_1=\text{\textit{Europe}})\rightarrow (o_1^*=\text{\textit{Africa}})$.
        This highlights the need for set-based operations instead of modifying a single object, as seen in previous approaches. 
        Our proposed formulation addresses this need by considering the KEO scenarios.

\begin{figure*}
    \centering
    \includegraphics[width=1\linewidth]{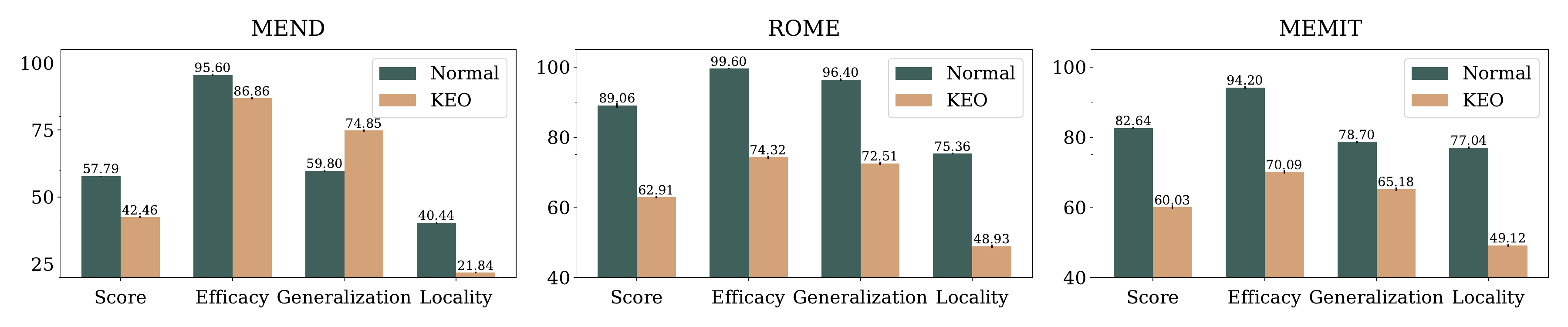}
    \caption{Comparison of editing performance on Normal and KEO type for MEND, ROME, and MEMIT.}
    \label{fig:norm_vs_keo}
\end{figure*}

    \paragraph{Evaluation Metrics}\label{sec:metrics}
        The evaluation metrics for the new formulation of KSE remain consistent with previous works (where the object is singular) \citep{meng2022locating,meng2022mass}. 
        These metrics assess the editing performance from the following perspectives:
        \textbf{Efficacy} 
            measures whether the post-edit model $f_{\theta}^*$ generates the intended predictions for the target edits, which is quantified using the Efficacy Score (ES); 
        \textbf{Generalization} 
            measures the ability of the post-edit model $f_{\theta}^*$ to generalize edits across equivalent inputs, which is evaluated via the Generalization Score (GS);
        \textbf{Locality} 
            measures whether the post-edit model $f_{\theta}^*$ retains its original predictions for inputs outside the editing scope (i.e., inputs that are not edited), which is evaluated by the Locality Score (LS).
        Higher scores for these metrics indicate better performance in their respective dimensions.
        Note that the metrics are evaluated based on the prompts provided in each instance.
        For detailed score calculations, please refer to Appendix \ref{app:metric}. 
\section{Pilot Analyses \& Dataset Construction}

\label{sec:pilot}

    We conduct pilot analyses on existing KE datasets to examine the prevalence of KEO instances and evaluate their impact on current KE methods. 
    Based on these findings, we construct a new KE dataset to further investigate the KEO problem.

    \subsection{KEO in Knowledge Editing}
        
        \paragraph{Statistics of Instances with KEO}
            
            We analyze and quantify the instances with KEO within the following KE datasets. 
            
            $\bullet$ \textbf{zsRE} \citep{levy2017zero} is one of the most prevalent QA datasets, extended and adopted by \citet{de2021editing,mitchell2021fast,mitchell2022memory} for KE evaluations.
            
            $\bullet$ \textbf{\textsc{ParaRel}} \citep{elazar2021measuring} is an expert-curated dataset containing diverse prompt templates for 38 relations sourced from the T-REx dataset \citep{elsahar2018t}.
            
            $\bullet$ \textbf{\textsc{MQuAKE-t}} and \textbf{\textsc{MQuAKE-cf}} \citep{zhong2023mquake} are constructed from Wikidata to evaluate the effectiveness of KE methods on multi-hop questions.
            
            $\bullet$ \textbf{\textsc{CounterFact}} 
            \citep{meng2022locating} originates from \textsc{ParaRel}, each sample includes a knowledge triplet and meticulously crafted prompt templates.

            The distribution of KEO in knowledge triplets within the aforementioned KE datasets is summarized in Table \ref{tab:keo_ratio}, which reveals the presence of KEO in all of them.
            Among the datasets analyzed, the zsRE dataset has the highest proportion of Normal triplets, accounting for 80.66\%. 
            In contrast, the Normal type constitutes only 7.29\% and 5.92\% in the \textsc{ParaRel} and \textsc{CounterFact} datasets, respectively. 
            Moreover, the \textsc{MQuAKE-t} and \textsc{MQuAKE-cf} datasets not only exhibit the KEO problem but also contain a substantial number of duplicate knowledge triplets, with sample sizes of 1772 and 3486, respectively.
            These statistics highlight the pervasive nature of knowledge element overlap in current mainstream KE datasets. 
            However, the practical implications of this issue have been insufficiently considered.
        
        \paragraph{Impact of KEO}\label{sec:keo}
        
            Given that previous editors have primarily been designed with a focus on the Normal type setting, neglecting the presence of KEO, we conduct an evaluation using 500 Normal triplets and 500 KEO triplets from the \textsc{CounterFact} dataset to investigate the impact of KEO on editors' performance. 
            As the experimental results shown in Figure \ref{fig:norm_vs_keo}, the performance of current SOTA editors generally exhibits lower effectiveness (except for the Generalization metric of MEND) when dealing with KEO-type knowledge compared to Normal type knowledge.
            Further analyses reveal that the performance discrepancy arises from the treatment of knowledge as key-value pairs stored in the models' feed-forward network (FFN). 
            When editing knowledge with element overlap, these approaches often lead to knowledge overwriting, wherein the value vector associated with the editing target is overwritten. 
            As a consequence, this results in decreased Efficacy and Generalization performance. 
            Moreover, the ripple effects of knowledge updates \citep{cohen2024evaluating} contribute to a decline in the Locality performance of instances with KEO.
            In summary, the observations above highlight the inadequacy of existing methods in handling KE with KEO, calling for a fresh perspective to address the challenges.

\begin{table}[tbp]
    \centering
    \caption{Statistics on current mainstream KE datasets.  Note that a knowledge triplet can belong to different classes. Duplicate (\textbf{Dup.}) is exact identical knowledge triplets. \textbf{Ratio} calculates the ratios of Normal (\textbf{Norm.}) instances in \textbf{ALL} instances.}
    \label{tab:keo_ratio}
    \resizebox{1.0\columnwidth}{!}{
    \begin{tabular}{lccccccc}
        \toprule
        \textbf{Dataset}     & \textbf{Norm.} & \textbf{RSO} & \textbf{ROO} & \textbf{SOO} & \textbf{Dup.} & \textbf{Ratio} & \textbf{ALL} \\ 
        \midrule
        zsRE                 & 8,066           & 0            & 902          & 1,103        & 0   & 80.66  & 10,000       \\
        \textsc{ParaRel}     & 2,023           & 2,242        & 23,241       & 742          & 21     &  7.29    & 27,738       \\
        \textsc{MQuAKE-t}    & 79              & 0            & 3            & 14           & 1,772   &   4.23    & 1,868        \\
        \textsc{MQuAKE-cf}   & 1,748           & 28           & 3,970        & 0            & 3,486    & 18.96    & 9,218        \\
        \textsc{CounterFact} & 592             & 315          & 9,376        & 10           & 11 & 5.92 & 10,000       \\ 
        \bottomrule
    \end{tabular}
    }
\end{table}

    \subsection{New Dataset: \textsc{EditSet}}\label{sec:con_data}

        To further explore improvements in KE for instances with KEO, we introduce a novel dataset, \textsc{EditSet}, designed exclusively with KEO samples to provide a focused evaluation environment.

        \paragraph{Data Collection}
            
            All knowledge triplets are collected from Wikidata,
            a knowledge base containing fact triplets associated with millions of entities.
            To collect KEO instances, we first sample subject-object pairs from Wikidata that share common relation properties.
            Subsequently, we utilize Wikidata dumps 
            to extract data, identifying 710 specific relations and retaining only instances where multiple objects are associated with a given factual statement $t_r(s)$.
            Using this method, knowledge triplets are collected.
            Finally, GPT-4 is employed to generate evaluation prompts, including counterfactual, paraphrase, and neighborhood prompts, based on the knowledge triplets (see Appendix \ref{app:prompt}  for prompt details).  %

        \paragraph{Data Statistics of \textsc{EditSet}}

            The \textsc{EditSet} dataset encompasses a total of 710 relations. 
            Additionally, we construct a subset comprising the 31 most frequently utilized relations that intersect with the \textsc{CounterFact} \citep{meng2022locating} and \textsc{ParaRel} \citep{elazar2021measuring} datasets. 
            In \textsc{EditSet}, each data sample is associated with $N$-objects, indicating that within each knowledge triplet (individual data instance), both the subject $s$ and relation $r$ are singular, while they correspond to $N$ objects $o$s.
            The subset comprises a total of 40,904 instances.
            Detailed statistics regarding the subset, including the total number of subjects and objects, are presented in Table \ref{tab:statistic}. 
            The counterfactual prompt is employed to assess \textbf{Efficacy}, the paraphrase prompt for \textbf{Generalization}, and the neighborhood prompt for \textbf{Locality}.
             
\begin{table}[htbp]
    \centering
    \caption{The statistics of \textsc{EditSet} dataset on 31 commonly used relations. The \textsc{EditSet} dataset consists of three types of prompt, Counter.P., Para.P., and Neigh.P. denote Counterfactual Prompt, Paraphrase Prompt, and Neighborhood Prompt, respectively. Each corresponds to a factual knowledge statement containing $N$-objects.}
    \label{tab:statistic}
    \resizebox{1.0\columnwidth}{!}{
    \begin{tabular}{lccccccc}
    \toprule
    \textbf{Overlap} & \textbf{N=3} & \textbf{N=4} & \textbf{N=5} & \textbf{N=6} & \textbf{N=7} & \textbf{N\textgreater{}=8} & \textbf{Total} \\
    \midrule
    Subjects         & 21,256       & 9,203        & 4,389        & 2,161        & 1,160        & 682                        & 35,301         \\
    Relations        & 31           & 28           & 28           & 22           & 22           & 19                         & 31             \\
    Objects          & 18,497       & 13,985       & 10,570       & 7,891        & 6,019        & 4,444                      & 26,144         \\
    Counter.P.       & 22,770       & 9,574        & 4,503        & 2,193        & 1,175        & 687                        & 40,900         \\
    Para.P.          & 43,164       & 18,410       & 8,674        & 4,219        & 2,249        & 1,325                      & 78,031         \\
    Neigh.P.         & 3,780        & 2,768        & 2,240        & 1,757        & 1,509        & 1,221                      & 3,988          \\
    \bottomrule
    \end{tabular}
    }
\end{table}
            The dataset is designed explicitly for KSE to support the exploration of KE on instances involving KEO.

\section{SetKE: A Set Knowledge Editor}

\begin{figure}[t]
    \begin{center}
    \includegraphics[width=1\linewidth]{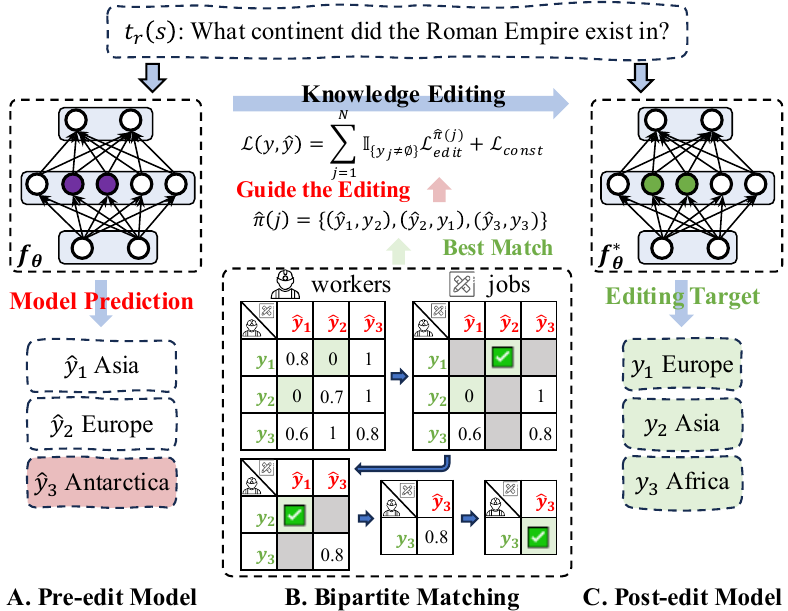}
    \end{center} \caption{ \label{fig:framework}
    Simplified illustration of the SetKE framework. 
    }
\end{figure}

    This section details the proposed Set Knowledge Editor (SetKE) tailored to the KEO issue.
    Traditional KE methods have not yet considered KEO, making them inadequate for addressing this issue, as empirically demonstrated in Section \ref{sec:pilot}. 
    In response, we depart from the patterns established by prior work in the KE task and approach the problem from an entirely new perspective. 
    Specifically, our goal in KSE is to transform a set $O$ into a new set $O^*$ to ensure the model is successfully edited. 
    This can be viewed as editing (or aligning) the model's prediction (the actual output set) to match the ground truth (post-edit outputs), establishing a mapping between the current set and the target set. 
    Inspired by the set prediction task \citep{sun2021rethinking}, we aim to constrain and optimize this mapping to achieve the desired transformation. 
    To this end, we employ bipartite graph matching to find the optimal correspondence, with the matching computed using the Hungarian algorithm. 
    The model weights are then updated based on the optimization, completing the KE process, and successfully editing the model.
    Figure \ref{fig:framework} shows a simplified framework for editing the model from a set perspective.

\subsection{Bipartite Matching Constraint} 
\label{sec:bipartite}

        Based on the background and considerations presented earlier, we approach the optimization of editing KEO triples using the bipartite matching problem. 
        In bipartite matching, given a set of predictions and a set of editing targets, the goal is to find the optimal matching between these two sets that maximizes the overall matching score while minimizing the total matching cost.
        Building on this basis, the KSE problem is defined as follows: 
        Given a set of editing targets $\mathbf{y} = \{y_j\}_{j=1}^M$ and a set of model predicted objects $\hat{\mathbf{y}} = \{\hat{y}_j\}_{j=1}^N$, where $M < N$, and placeholders ($\emptyset$) are used to pad unmatched predictions, the task is to find an optimal matching $\hat{\pi}$ that minimize the total matching cost, determined by:
        \begin{equation}
            \hat{\pi} = \arg \min_{\pi \in \Pi_N} \sum_{j=1}^N C_{\text{match}}(y_j, \hat{y}_{\pi(j)}),
        \end{equation}
        where $\hat{y}_{\pi(j)}$ is the matching of $y_j$ under the match $\pi$, $\Pi_N $ represents the set of all matchings.
        $C_{\text{match}}(y_j, \hat{y}_{\pi(j)})$ is the pairwise matching cost, which, in this work, is defined as:
        \begin{equation}
            C_{\text{match}}(y_j, \hat{y}_{\pi(j)}) = - \mathds{1}_{\{y_j \neq \emptyset\}} P_{\pi(j)}(y_j),
        \end{equation}
        where $P_{\pi(j)}(y_j)$ represents the probability that the model's predicted object $\hat{y}_{\pi(j)}$ correctly matches the editing target $y_j$.

        Building on this bipartite matching problem, we employ the Hungarian algorithm \citep{kuhn1955hungarian} to find the optimal matching. 
        Figure \ref{fig:framework}B illustrates a toy example of optimal matching obtained using the Hungarian algorithm.
        The algorithm treats the set of ground truth objects $\{y_j\}_{j=1}^M$ as a set of workers and the set of predicted objects $\{\hat{y}_j\}_{j=1}^N$ as a set of jobs. 
        Given a matrix $\mathbf{A}^{N \times N}$ padded with $0$, each element $A_{ij}$ represents the cost of assigning job $j$ to the worker $i$.
            To begin, the algorithm reduces the matrix by first subtracting the minimum value in each row from all elements in that row, ensuring that the smallest cost for each worker is minimized. 
            Then, it proceeds by subtracting the minimum value in each column from all elements in that column, further simplifying the matrix and reducing the overall cost values.
            Next, the elements that are now zero represent potential optimal assignments. 
            The algorithm assigns these zeros to form part of the optimal matching and iteratively adjusts the matrix until all workers are assigned jobs that give the minimal total cost.
        The Hungarian algorithm guarantees finding the optimal matching in $O(N^3)$ time complexity, as shown in Appendix Algorithm 1.  

        Once the optimal matching (or assignment) $\hat{\pi}$ is found, we define the overall loss as: 
        \begin{equation}
            \mathcal{L}_{\text{}}(\mathbf{y}, \hat{\mathbf{y}}) = \sum_{j=1}^N  \mathds{1}_{\{y_j \neq \emptyset\}} \mathcal{L}_{\text{edit}}^{ \hat{\pi}(j)}+\mathcal{L}_{\text{const}},
        \label{eq:total_opt}
        \end{equation}
        where $\mathcal{L}_{\text{edit}}^{\hat{\pi}(j)}$ is the edit objective loss that measures the success of the edit, and  $\mathcal{L}_{\text{const}}$ is the constraint loss that optimizes the edit locality metric. The overall process is computed for each ground truth $y_j$ and its corresponding prediction $\hat{y}_{\hat{\pi}(j)}$, as follows:
        \begin{equation}
        \begin{aligned}
            \mathcal{L}_{\text{edit}}^{ \hat{\pi}(j)} &= - \log P_{\theta(h_i^l + \delta_i)} \left[ y_{ij} \mid x_i \right]_{j = { \hat{\pi}(j)}}, \\
            \mathcal{L}_{\text{const}} &= D_\mathrm{KL}\big(P_{\theta(h_i^l + \delta_i)} \left[ y^{\prime} \mid x^{\prime} \right] \| P_\theta \left[ y^{\prime} \mid x^{\prime} \right]\big).
        \end{aligned}
        \end{equation}
        The second term $\mathcal{L}_{\text{const}}$  minimizes the KL divergence of predictions for the input $x^{\prime}$ (of the form “{subject} is a”) to the unchanged model, which helps preserve the model’s understanding of the subject’s essence.
        Intuitively, $\mathcal{L}_{\text{edit}}$ is small when the model successfully updates its output, while $\mathcal{L}_{\text{const}}$ is small when the edit does not affect unrelated inputs.

    \subsection{Model Weight Updating for KE}

        Existing KE methods \citep{meng2022mass,li2024pmet} apply optimization exclusively to specific parts of the model (those containing the knowledge to be edited) to avoid affecting non-edit target. 
        This work follows this \textit{locate-then-edit} paradigm. 
        Based on the optimization strategy introduced in Section \ref{sec:bipartite}, the editing process proceeds as follows:

        \textbf{\textit{Step 1: Locating the Editing Component.}}
            Current KE works \citep{meng2022locating,li2024pmet} have shown that knowledge generally resides in specific model layers, such as the feed-forward network (FFN) in Transformer-based models. 
            Therefore, editing typically occurs within these layers. 
            In this step, we build on previous works \citep{meng2022locating} to trace the location of the component to be edited.
        
        \textbf{\textit{Step 2: Perform the Edit.}}
            Given the traced editing component, specifically layer $l$ in the model $f_{\theta}$, where the editing is performed, and a given editing prompt $x_i$ (describing the subject and relation of the set of KEO triplets $(s_i, r_i)$), we compute the hidden state $h_i^l$ after the prompt passes through the $l$-th FFN layer, as follows:
            \begin{equation}
                h_i^l = \operatorname{FFN}^l(x_i) = \sigma(x_i \cdot W_{fc}^l) \cdot W_{proj}^l.
            \end{equation}
        %
            Since the goal is to edit specific knowledge without affecting other parts of the model, we introduce a residual vector $\delta_i$ to adjust the model’s output, using this vector to perform the optimization.
            This residual enables the transfer of knowledge from the original value $v_i^l$ to the edited value $z_i$, thereby achieving the desired edit:
            \begin{equation}
                z_i = v_i^l + \delta_i = h_i^l + \arg\min_{\delta_i} \mathcal{L}(\delta_i).
            \end{equation}

        \textbf{\textit{Step 3: Spreading Edits Across Layers.}}  
            To minimize side effects—such as unintended modifications or overwriting—and reduce excessive modifications, we distribute the residual vector $\delta_i$ across multiple layers of the model, rather than applying it to a single layer. 
            This is achieved following prior works \citep{meng2022mass, li2024pmet}, which evenly spreads the updates across critical layers:
            \begin{equation}
                r_i^l = \frac{\delta_i}{L - l + 1}, \quad \mathcal{R}^l \triangleq [r_1^l \mid r_2^l \mid \cdots \mid r_n^l].
            \end{equation}
        
        \textbf{\textit{Step 4: Weight Update.}}
            Finally, the model’s weight matrix $W_{proj}^l$ is updated by adding the incremental weight $\Delta^l$ to the original weight, as computed below:
            \begin{equation}
                \Delta^l = \mathcal{R}^l K^{l \top} (C^l + K^l K^{l \top})^{-1},
            \end{equation}
            where $\hat{W}^l_{proj} = W^l_{proj} + \Delta^l$ represents the updated weights, $K^l$ represents the new keys, $C^l = K_0^l {K_0^l}^\top$ is an estimate of the set of previous keys obtained through sampling, and $\mathcal{R}^l$ is the residual between the old and new value vectors.
\section{Experiments}

\begin{table*}[htbp]
    \centering
    \caption{\label{tab:main_result} Numerical results on \textsc{EditSet} for 10,000 edits (95\% confidence intervals in parentheses).}
    \begin{tabular}{lcccc|cccc}
    \toprule
    \multicolumn{5}{c}{\textbf{GPT2 Large (760M)}} & \multicolumn{4}{c}{\textbf{GPT2 XL (1.5B)}} \\
    \midrule
    \multirow{1}{*}{\textbf{Editor}} & \textbf{Score } & \textbf{Efficacy } & \textbf{Generalization } & \textbf{Locality } & \textbf{Score } & \textbf{Efficacy } & \textbf{Generalization } & \textbf{Locality } \\
     \midrule
    FT-W & 51.46 & 58.13 (0.5) & 45.29 (0.5) & 52.58 (0.4) & 55.18 & 65.66 (0.5) & 50.97 (0.5) & 51.24 (0.4) \\
    KN & 42.21 & 38.22 (0.5) & 37.46 (0.5) & 54.89 (0.4) & 40.65 & 35.93 (0.5) & 35.86 (0.5) & 55.29 (0.4) \\
    MEND & -- & -- & -- & -- & 38.75 & 89.58 (0.3) & 81.58 (0.4) & 18.52 (0.3) \\
    PMET & 43.34 & 40.82 (0.5) & 39.24 (0.5) & 51.98 (0.4) & 44.13 & 41.15 (0.5) & 40.08 (0.5) & 53.39 (0.4) \\
    MEMIT & 49.65 & 49.53 (0.5) & 45.64 (0.5) & 54.59 (0.4) & 56.60 & 61.12 (0.5) & 55.07 (0.5) & 54.11 (0.4) \\
    ROME & 64.08 & 72.03 (0.4) & 69.63 (0.4) & 53.84 (0.4) & 65.89 & 75.27 (0.4) & 73.58 (0.4) & 53.62 (0.4) \\
     \midrule
    SetKE  & {71.47} & {88.57 (0.3)} & {80.91 (0.4)} & {54.77 (0.4)} & {75.28} & {95.90 (0.2)} & {91.68 (0.2)} & {54.14 (0.4)} \\
    \bottomrule
    \end{tabular}
    
\end{table*}

In this section, we evaluate the editing performance of baselines and SetKE on \textsc{EditSet} concerning KEO knowledge.
Specifically, our goal is to answer the following questions:
\textbf{Q1}: How does SetKE perform compared to other baselines on \textsc{EditSet}?
\textbf{Q2}: How does the quantity of KEO overlaps affect the knowledge editing methods?
\textbf{Q3}: How does the KEO phenomenon lead to the problem of knowledge overwriting?
\textbf{Q4}:  
How does the bipartite matching constraint contribute to the results?

    \subsection{Experimental Settings}

        \paragraph{Dataset}

            The editing is performed and evaluated using a subset of the proposed dataset, \textsc{EditSet}, which comprises 31 commonly used relations.
            For specific statistics and detailed configurations of this subset, please refer to Section \ref{sec:con_data}.
            In this dataset, all KEO types are RSO types and each instance is represented as $\{s,r,O=\{o_1, o_2, ...\}\}$, aligning with the objective of editing instances where the object is not singular.

        \paragraph{Language Models} 
            
            We employ two widely adopted autoregressive language models, namely GPT2-Large (760M), GPT2-XL (1.5B)  and GPT-J (6B) \citep{radford2019language}, as the base language models to perform editing and assess the effectiveness of the KE approaches.

        \paragraph{Baselines} 
            We select the following approaches:
            \textbf{FT-W} is a basic fine-tuning method.
            \textbf{KN} \citep{dai2022knowledge} utilizes knowledge attribution to implement knowledge updates.
            \textbf{MEND} \citep{mitchell2021fast} uses low-rank decomposition of gradients to learn new knowledge.
            \textbf{ROME} \citep{meng2022locating} first applies causal mediation analysis to updates the parameters.
            \textbf{MEMIT} \citep{meng2022mass} extends ROME to edit a large batch of facts. 
            \textbf{PMET} \citep{li2024pmet} extends MEMIT to optimize the hidden states of both the FFN and Attention modules simultaneously. 

    \subsection{Main Results (Q1)}
    
        Table \ref{tab:main_result} showcases numerical results on GPT2-Large (760M) and GPT2-XL (1.5B) over 10,000 cases in \textsc{EditSet}, respectively. 
        In this experiment, we compare with the recent baselines and our proposed method SetKE. 
        We observe that SetKE is significantly superior to the exiting editing methods without using any additional parameters.
        Specifically, we outperform the recent
        advanced batch editing baseline MEMIT by up to 39.04\% and  35.27\% improvements regarding to the Efficacy and Generalization metrics in most cases for GPT2-Large.
        Under the same experimental conditions on GPT2-XL, SetKE also achieves a better performance improvement of 9.39\%, 20.63\% and 18.1\% regarding to the Score, Efficacy and Generalization metrics, respectively, relative to the state-of-the-art methods.
        which indicates the effectiveness of our method in accurately altering models’ behavior for the editing overlap factual knowledge without interference on each other.

\subsection{Analysis for Overlap Quantity (Q2)}
\begin{figure}[h]
   \begin{center}
   \includegraphics[width=1 \linewidth]{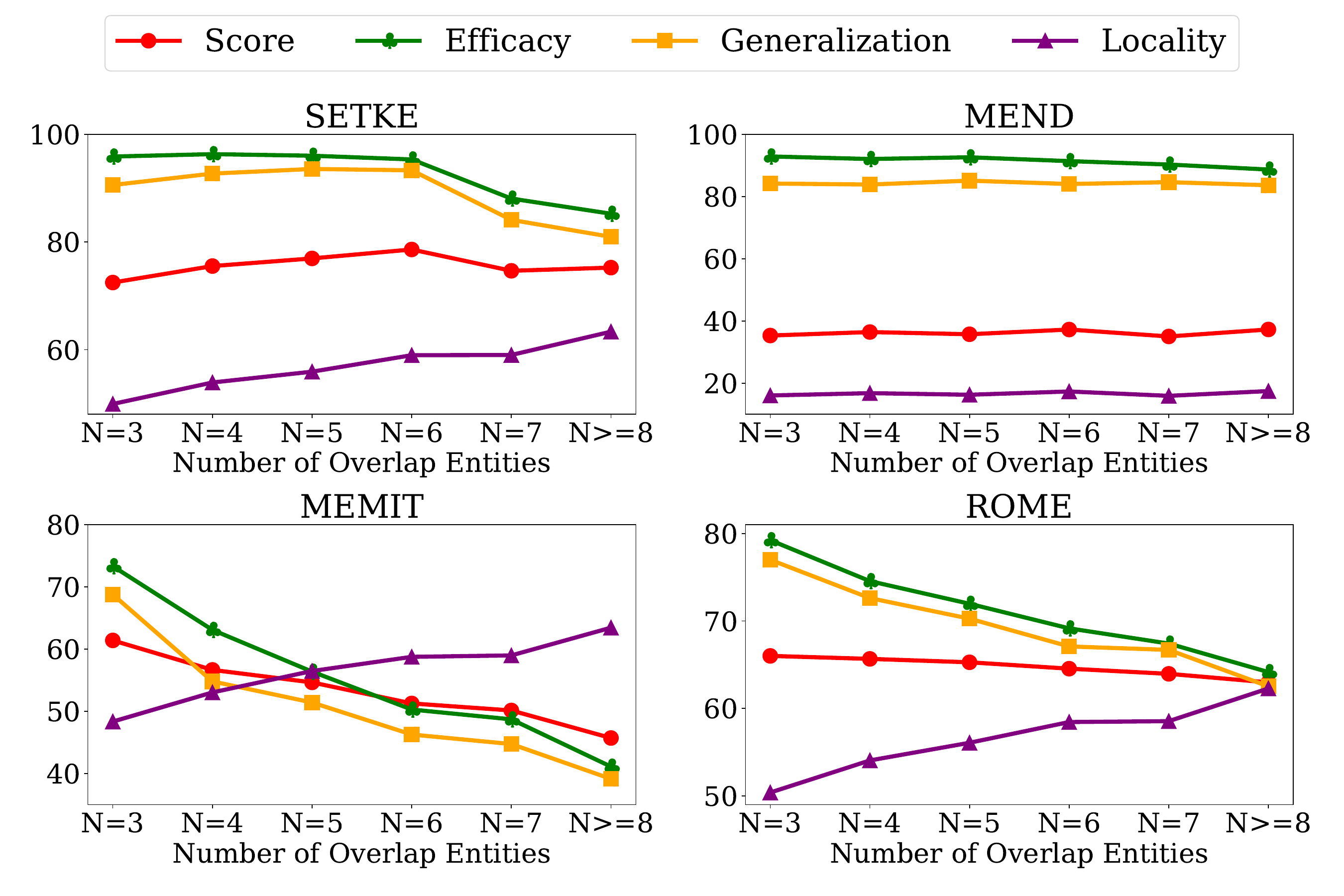}
   \end{center}
   \caption{ \label{fig:overlap_entity_count}
   Comparing the impact of knowledge overlap number on baselines and SetKE.
   }
\end{figure}

Figure \ref{fig:overlap_entity_count} illustrates how the performance of editors changed with respect to different numbers of knowledge  triplets overlap on \textsc{EditSet} with GPT2-XL edited by different methods.
We analyze a total of 3000 KEO cases and 
it can be seen from these results that the performance of all editing methods degrades as the number of overlap triplets increased.
Figure \ref{fig:overlap_entity_count} illustrates the performance variation of editors concerning different levels of overlap in knowledge triplets within \textsc{EditSet} edited using GPT2-XL by distinct methods. 
The results indicate a general decrease in performance across all editing techniques as the number of overlapping triplets increases.
Specifically, ROME and MEMIT demonstrate heightened sensitivity to the KEO problem, exhibiting a significant decline in editing performance. On the other hand, MEND shows relatively less susceptibility to KEO's impact. This disparity can be attributed to the knowledge localization approach of ROME and MEMIT, which tends to lead to more instances of knowledge overwriting in KEO scenarios.
Furthermore, while SetKE also operates within the realm of knowledge localization, it exhibits lesser vulnerability compared to ROME and MEMIT. This suggests that the bipartite matching constraint in SetKE enhances editing performance in KEO cases by minimizing the occurrence of knowledge overwriting.
Finally, the Locality metric continues to improve across all methods as the number of overlaps increases. We attribute this trend to the overfitting of $\mathcal{L}_{\text{const}}$.

\begin{figure*}[ht]
   \begin{center}
   \includegraphics[width=1\linewidth]{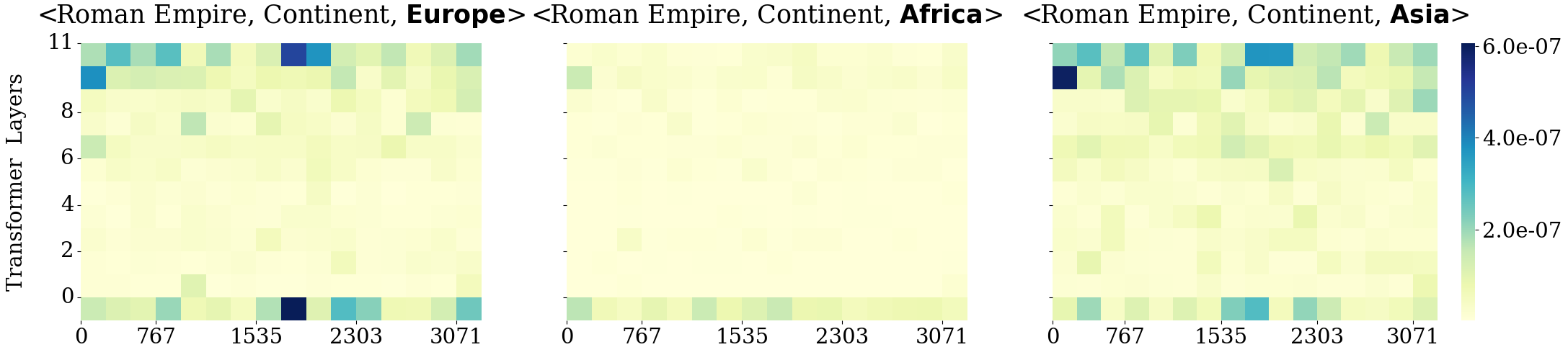}
   \end{center}
   \caption{\label{fig:heatmap}
   The result of knowledge localization of KEO type knowledge on GPT2.
   }
\end{figure*}

\begin{table*}[htbp]
    \centering
    \caption{\label{tab:result_gptj} The results of GPT-J (6B) on \textsc{EditSet} for 7,500 edits (95\% confidence intervals in parentheses).}
    \begin{tabular}{lcccc|cccc}
    \toprule
    \multicolumn{5}{c}{\textbf{Object Set}} & \multicolumn{4}{c}{\textbf{Object Concatenation}} \\
    \midrule
    \multirow{1}{*}{\textbf{Editor}} & \textbf{Score } & \textbf{Efficacy } & \textbf{Generalization } & \textbf{Locality } & \textbf{Score } & \textbf{Efficacy } & \textbf{Generalization } & \textbf{Locality } \\
    \midrule
    FT-W & 46.65 & 51.62 (0.5) & 39.92 (0.5) & 50.29 (0.4) & 43.36 & 44.53 (0.5) & 37.20 (0.5) & \textbf{50.40 (0.4)} \\
    MEND & 37.44 & \textbf{91.33 (0.3)} & 82.60 (0.4) & 17.52 (0.3) & 43.05 & 67.64 (0.5) & 63.37 (0.5) & 25.56 (0.4) \\
    MEMIT & 55.95 & 67.51 (0.5) & 56.04 (0.5) & 47.71 (0.4) & 57.07 & 82.31 (0.4) & 64.00 (0.4) & 40.34 (0.4) \\
    ROME & 59.48 & 80.54 (0.4) & 79.90 (0.4) & 39.21 (0.4) &\textbf{ 61.30} & 85.86 (0.3) & \textbf{82.57 (0.3)} & 39.71 (0.4) \\
    \midrule
    SetKE & \textbf{73.68} & 90.08 (0.3) & \textbf{90.76 (0.3)} & \textbf{53.77 (0.4)} & 59.50 & \textbf{89.43 (0.3)} & 68.68 (0.4) & 40.52 (0.4) \\
    \bottomrule
    \end{tabular}
\end{table*}

\subsection{Analysis for Knowledge Overwriting (Q3)}
In this study, we conduct an analysis of 500 KEO cases to investigate whether KEO knowledge triplets could lead to the problem of knowledge overwriting, as shown in Figure \ref{fig:showcase}(b). Following the knowledge attribution method \citep{dai2022knowledge}, which is based on integrated gradients to determine the storage location of specific factual knowledge within the model (knowledge neurons), we analyze the degree of overlap among the knowledge neurons corresponding to KEO knowledge to assess whether knowledge triplets associated with KEO could potentially trigger knowledge overwriting.
Our analysis results, as shown in Figure \ref{fig:heatmap} for a representative example, reveal a significant overlap among the knowledge neurons representing three KEO factual pieces of knowledge in the GPT2 \citep{radford2019language} model. This observation aligns with our pilot analysis (see Figure \ref{fig:norm_vs_keo}), indicating that repetitive modifications to these specific knowledge neurons often lead to knowledge overwriting, resulting in a decline in editing performance.

\subsection{Analysis for Ablation Study (Q4)}

To evaluate the effectiveness of SetKE optimized through the bipartite matching constraint, we compared it with vanilla knowledge set editing (object concatenation). This setting treats the editing target as a long sequence of multiple concatenated objects, aiming to optimize multiple objects simultaneously.
The results are displayed in Table \ref{tab:result_gptj}. The left section of the table (object set) showcases the editors' performance aligned with the primary experimental configuration, while the right side illustrates the outcomes of the vanilla (object concatenation) setup. In the KEO scenario, we observe that optimizing multiple objects simultaneously generally outperforms iterative object optimization for editors such as MEND, ROME, and MEMIT. We speculate that this is because optimizing multiple objects simultaneously helps mitigate the issue of knowledge overwriting.
Under object concatenation setting, SetKE exhibits lower performance compared to ROME. Nevertheless, all editors fall short of SetKE's performance in the object set configuration, highlighting the effectiveness of the bipartite matching constraint specifically designed for SetKE in the set scenario.

\section{Related Work}\label{sec:bibtex}

Knowledge Editing (KE) involves modifying language models to revise the expression of factual knowledge learned from the pre-training corpus. 
The current work on the KE task can be categorized into the following three types:
Meta-learning methods utilize additional trainable parameters to store memory or learn the required adjustments ($\Delta$) to update knowledge in LLM \citep{de2021editing,huang2022transformer, tan23malmen, cheng2024editing}.
Locate-Then-Edit methods first employ causal mediation analysis to locate knowledge neurons that exhibit a positive correlation with a knowledge expression, and then modify them accordingly \citep{dai2022knowledge, meng2022locating, meng2022mass,huang2024reasons, huang2024commonsense, wang2024editing}.
In-Context Edit methods are a training-free paradigm where knowledge editing is achieved directly by concatenating demonstrations within the input context \citep{zheng2023can, zhong2023mquake, qi2024context}. 
However, existing KE methods are primarily restricted to modifying standard triplets and often overlook the phenomenon of knowledge overlap. To the best of our knowledge, this paper is the first to introduce a new dataset, \textsc{EditSet}, designed to evaluate KEO editing performance, and proposes a general framework to mitigate knowledge overwriting.

\section{Conclusion}

    In this paper, we analyze mainstream knowledge editing (KE) datasets and identify a widespread yet previously overlooked issue: the Knowledge Element Overlap (KEO) phenomenon. 
    We also demonstrate that existing editing methods are insufficient to effectively address this challenge. 
    To tackle this limitation, we propose a novel formulation, Knowledge Set Editing (KSE), and introduce SetKE, a method specifically designed for KEO scenarios.
    SetKE leverages bipartite matching to optimize object set editing, effectively resolving editing conflicts and improving accuracy. 
    Experimental results confirm that SetKE outperforms existing methods, achieving state-of-the-art performance in editing multiple mainstream LLMs. 
    Furthermore, we develop the \textsc{EditSet} dataset, which serves as a comprehensive benchmark for evaluating knowledge set editing in KEO scenarios.

\bibliographystyle{named}
\bibliography{ijcai25}

\appendix
\newpage
\section{Appendix}
\label{sec:appendix}
\label{app:pilot_exp}

\subsection{Generating Questions using GPT-4}
\label{app:prompt}
we collect factual knowledge in the form of triples from Wikidata RSO. An example of this can be found at the provided link \footnote{\url{https://www.wikidata.org/wiki/Property_talk:P1302}}. 
We have a Property ID (Relation ID) and the corresponding relation description from Wikidata for this relation. 
Using this information, we prompt GPT-4 to automatically generate questions based on the description of the Property ID and the associated entity pairs. The prompt used for this purpose is presented in Table  \ref{tab:gpteneration} and Table \ref{tab:prompt_example}.
And, we show a sample of \textsc{EditSet} as show in Figure \ref{fig:case}.

\begin{figure}[h]
   \begin{center}
   \includegraphics[width=1\linewidth]{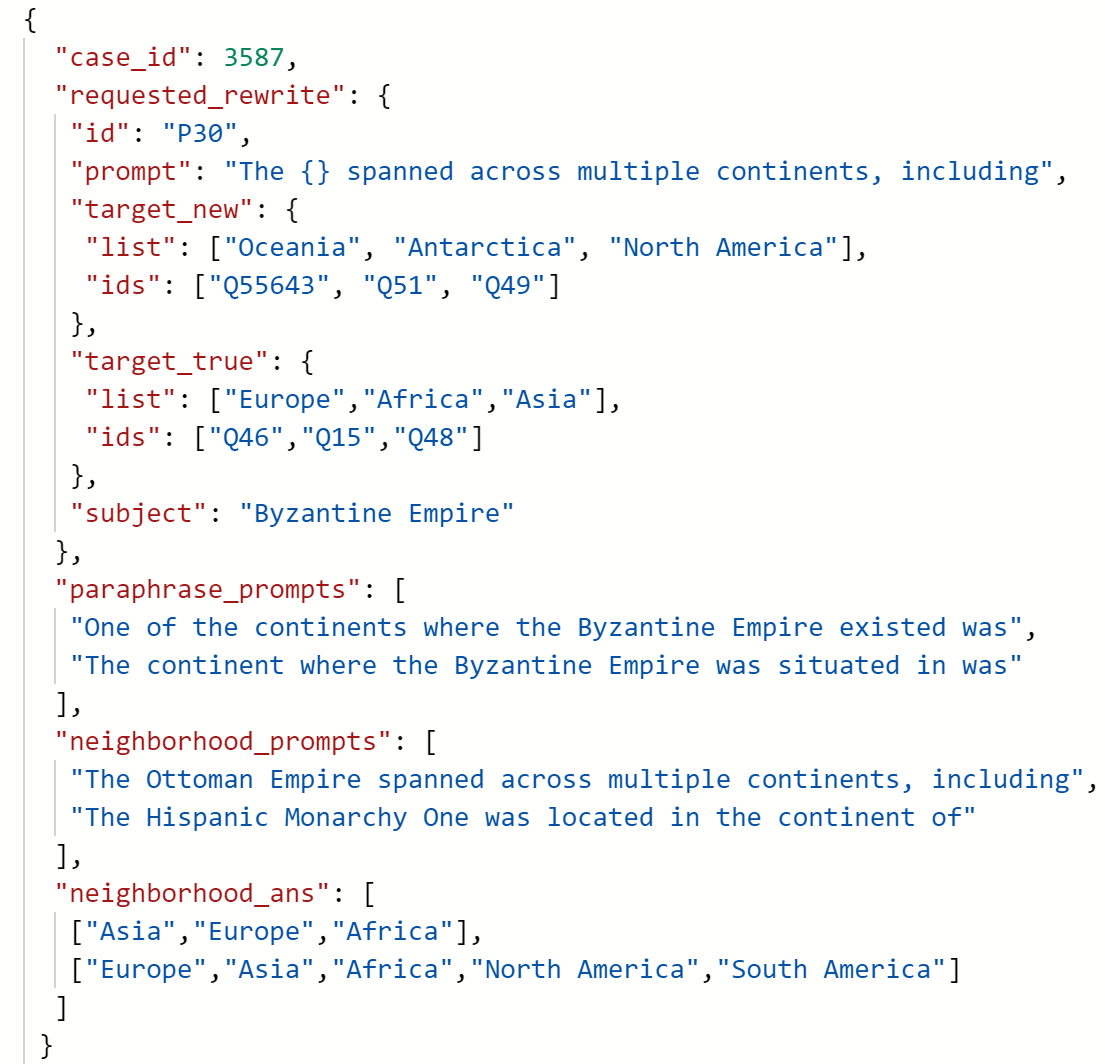}
   \end{center}
   \caption{\label{fig:case}
   Sample form of the \textsc{EditSet} dataset.
   }
\end{figure}

\begin{algorithm}[htbp]
            \renewcommand{\algorithmicrequire}{\textbf{Input:}}
            \renewcommand{\algorithmicensure}{\textbf{Output:}}
            \caption{Hungarian Algorithm}
            \label{alg:hungarian}
            \begin{algorithmic}[1] 
            \REQUIRE A cost matrix \( C \in \mathbb{R}^{N \times N} \)
            \ENSURE $\hat{\pi}$ 
            \STATE \textbf{Step 1:} Subtract row minima:
            \FOR{each row \( i \) in \( C \)}
            \STATE \( C[i, :] = C[i, :] - \min(C[i, :]) \)
            \ENDFOR
            \STATE \textbf{Step 2:} Subtract column minima:
            \FOR{each column \( j \) in \( C \)}
            \STATE \( C[:, j] = C[:, j] - \min(C[:, j]) \)
            \ENDFOR
            \STATE \textbf{Step 3:} \textbf{repeat}
            \STATE Cover all zeros with a minimum number of horizontal and vertical lines
            \STATE Find the smallest entry not covered by any line
            \STATE Subtract this entry from all uncovered entries
            \STATE Add this entry to all entries covered twice
            \STATE \textbf{until} the number of lines equals \( N \)
            \STATE \textbf{Step 4:} Find an optimal assignment among the zeros using DFS or BFS
            \STATE \textbf{return}  ~ $\hat{\pi}$
            \end{algorithmic}
        \end{algorithm}

\subsection{Implementation Details}
        The experiments are implemented using the PyTorch framework and run on a machine with eight NVIDIA GeForce RTX 3090 GPUs.
        We randomly sample 50\% and 50\% for training and testing on knowledge editing task.
        For evaluation, we follow the ROME   as the editing metrics.
        We set the batch size to triplet overlap number for each KEO case, the learning rate to be $5e-1$ and employed the Adam optimizer.
        We report the average and standard deviation of 5 repetitions.
        All implementations are available at \url{https://anonymous.4open.science/r/SetKE-11B4}.

The optimal assignment with the minimum total cost is easy to be computed via the  Hungarian algorithm (shown in Algorithm 1).
\subsection{Hungarian Algorithm }
The optimal assignment with the minimum total cost is easy to be computed via the  Hungarian algorithm (shown in Algorithm 1).

\subsection{Default Location Settings}

With editing tasks based on different LLM backbones, the
location of layer for SetKE can be treated as hyper-parameters. Following tables demonstrate the default location settings in our experiments (Table \ref{tab:large_loc} for GPT2-Large,
Table \ref{tab:xl_loc} for GPT2-XL).



\begin{table}[h]
\centering
\caption{\label{tab:large_loc}Default SetKE Target Modules for GPT2 Large}
\begin{tabular}{lc}
\hline
\textbf{Model} & \textbf{Target Editing Modules} \\
\hline
\multirow{5}{*}{GPT2 Large} & transformer.h.[1].mlp.c\_proj \\
 & transformer.h.[2].mlp.c\_proj \\
 & transformer.h.[3].mlp.c\_proj \\
 & transformer.h.[4].mlp.c\_proj \\
 & transformer.h.[5].mlp.c\_proj \\
\hline
\end{tabular}
\end{table}

\begin{table}[h]
\centering
\caption{\label{tab:xl_loc}Default SetKE Target Modules for GPT2 XL}
\begin{tabular}{lc}
\hline
\textbf{Model} & \textbf{Target Editing Modules} \\
\hline
\multirow{5}{*}{GPT2 XL} & transformer.h.[13].mlp.c\_proj \\
 & transformer.h.[14].mlp.c\_proj \\
 & transformer.h.[15].mlp.c\_proj \\
 & transformer.h.[16].mlp.c\_proj \\
 & transformer.h.[17].mlp.c\_proj \\
\hline
\end{tabular}
\end{table}

\subsection{Evaluation Metric}
\label{app:metric}




We compile a collection of challenging false statements $(s, r, o^*)$. 
These hypothetical scenarios initially receive lower rankings compared to the correct statements $(s, r, o)$. 
Our Efficacy Score (ES) represents the proportion of cases where we observe $P[o^*] > P[o]$ following editing. 
To evaluate generalization, we generate alternate prompts equivalent to $(s, r)$ for each counterfactual and calculate Generalization Scores (GS) in a manner akin to ES. 
For assessing Locality, we assemble a group of related subjects $s_n$ where $(s_n, r, o)$ remains valid.
\begin{itemize}
    \item \textbf{Efficacy Score (ES)} is the proportion of cases where \( o_i^* \) exceeds \( o_i \) in probability. Note that the prompt matches exactly what the edit method sees at runtime:
    \[
        \mathbb{E}_i \left[ \mathbb{P} \left[ o_i^* \mid p(s_i, r_i) \right] > \mathbb{P} \left[ o_i \mid p(s_i, r_i) \right] \right]. \tag{21}
    \]

    \item \textbf{Generalization Score (GS)} is the proportion of cases where \( o_i^* \) exceeds \( o_i \) in probability on rephrasings of the original statement:
    \[
        \mathbb{E}_i \left[ \mathbb{E}_{p \in \text{paraphrases}(s_i, r_i)} \left[ \mathbb{P} \left[ o_i^* \mid p \right] > \mathbb{P} \left[ o_i \mid p \right] \right] \right]. \tag{22}
    \]

    \item \textbf{Locality Score (LS)} is the proportion of neighborhood prompts where the models assign higher probability to the correct fact:
    \[
        \mathbb{E}_i \left[ \mathbb{E}_{p \in \text{neighborhood}(s_i, r_i)} \left[ \mathbb{P} \left[ o_i^* \mid p \right] < \mathbb{P} \left[ o_i \mid p \right] \right] \right]. \tag{23}
    \]

\end{itemize}

To explore the trade-off between generalization and locality, we introduce a composite measure, Editing Score (S), which is the harmonic mean of ES, GS, and LS.
When dealing with multiple objects, we utilize the arithmetic mean of editing scores to represent the performance of set editing.

\subsection{Performance on \textsc{CounterFact}}

Table \ref{tab:result_mcf} showcases numerical results on GPT2-XL (1.5B)  over 3,000 cases on \textsc{CounterFact} dataset.
In this experiment, we compare with the recent baselines and SetKE. 
We observe that SetKE also achieved the best performance on previous knowledge editing data.
\begin{table}[h]
\centering
    \caption{\label{tab:result_mcf} The results of GPT2-XL (1.5B) on \textsc{CountFact} for 3,000 edits (95\% confidence intervals in parentheses).}
\begin{tabular}{lcccc}
\toprule
Editor & Score & ES    & GS   & LS          \\ \midrule
FT     & 36.82 & 28.30 (0.4) & 30.97 (0.4) & 72.17 (0.3) \\
KN     & 29.76 & 21.57 (0.4) & 24.02 (0.3) & 78.06 (0.3) \\
MEND   & 41.26 & 35.27 (0.5) & 34.40 (0.4) & 65.38 (0.4) \\
ROME   & 53.57 & 56.67 (0.5) & 52.08 (0.4) & 52.22 (0.3) \\
PMET   & 46.19 & 44.33 (0.5) & 36.47 (0.4) & 66.85 (0.3) \\
\midrule
SETKE  & 66.58 & 70.53 (0.5) & 56.38 (0.4) & 76.08 (0.3) \\
\bottomrule
\end{tabular}
\end{table}


\begin{table*}[htbp]
\centering
\caption{\label{tab:gpteneration}
An example of questions generated by GPT-4 (gpt-4-turbo). 
We use python scripts to filter out problem templates that don't fit the format.}
\begin{tabular}{|l|}
\hline
\begin{tabular}[c]{@{}l@{}}
The Generation of ChatGPT : 
\\ 
\{ "template":  "The \{\} extended its influence to multiple continents, including" \}
\\
\{ "template":  "The \{\} spanned across multiple continents, including" \} 
\\
\{ "template":  "One of the continents where the \{\} existed was" \} 
\\
\{ "template":  "The continent where the \{\} was situated in was" \}
\\
\{ "template":  "The \{\} was located in the continent of" \}\\
\{ "template":  "The \{\} had a presence in various continents, such as"\}
\end{tabular} 
\\ \hline
\end{tabular}
\end{table*}

\begin{table*}[htbp]
\centering
\caption{\label{tab:prompt_example}
An example of using GPT-4 (gpt-4-turbo) to generate questions from Wikidata triples. We manually
write 3 demonstrations as the prompt when querying ChatGPT.}
\begin{tabular}{|l|}
\hline
\begin{tabular}[c]{@{}l@{}}Instruction: Please help me generate 10 templates by imitating the following example:\\ \\ Input:\\ subject = 'Roman Empire'\\  relation = 'continent'\\  object = {[}'Asia','Europe','Africa'{]}\\ \\  Output:\\  template\_list={[}'The Roman Empire spanned across multiple continents, including',\\ 'The Roman Empire was situated in the continents of',{]}\\  \\ In this context, the relation string must appear in the template. \\ The predicate in template is generated by relation, \\ and must satisfy all objects at the same time. \\ And the elements left blank at the end correspond to the object: \\ Asia, Europe, Africa, which cannot appear in template.\end{tabular} \\ \hline
\end{tabular}
\end{table*}

\subsection{Heatmap of Neuron Activations}
We aim to explore the correlation between knowledge overwriting issues and KEO triplets. To achieve this, we apply the knowledge attribution method \citep{dai2022knowledge} to GPT-2. 
Heatmaps are used to visualize neuron activation values, where darker colors represent higher activation values. The x-axis corresponds to the index of neurons, and the y-axis corresponds to their respective layers.

Through this approach, we aim to identify the positions within the model where knowledge is stored. Specifically, we find high overlap in the activation of neurons related to KEO triplets, whereas neurons involved in non-KEO instances do not exhibit strong correlations.

\begin{figure*}[h]
   \begin{center}
   \includegraphics[width=1\linewidth]{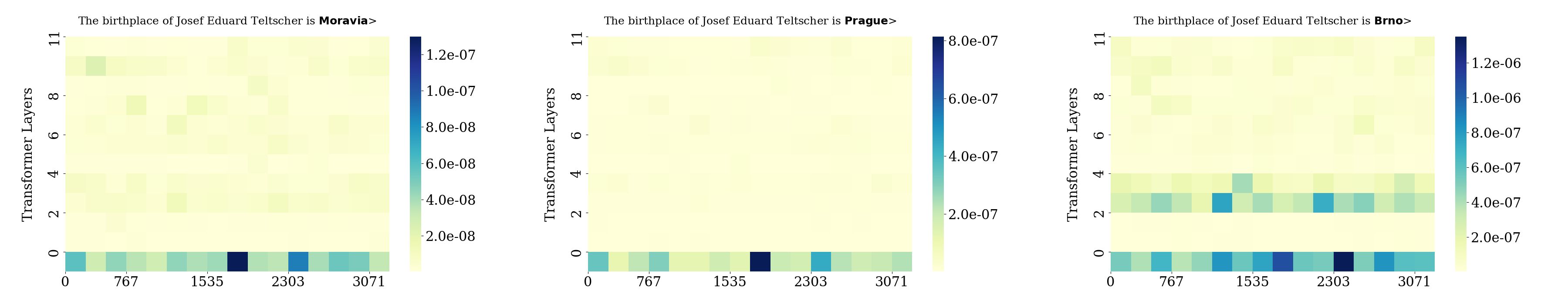}
   \end{center}
   \caption{\label{fig:case}
   Analysis for knowledge overwriting using case 1.
   }
\end{figure*}

\begin{figure*}[h]
   \begin{center}
   \includegraphics[width=1\linewidth]{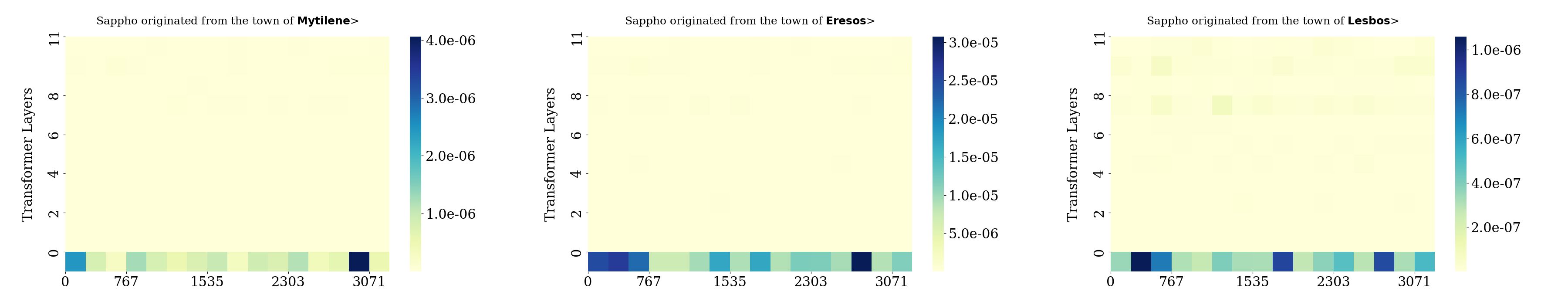}
   \end{center}
   \caption{\label{fig:case}
   Analysis for knowledge overwriting using case 2.
   }
\end{figure*}

\begin{figure*}[h]
   \begin{center}
   \includegraphics[width=1\linewidth]{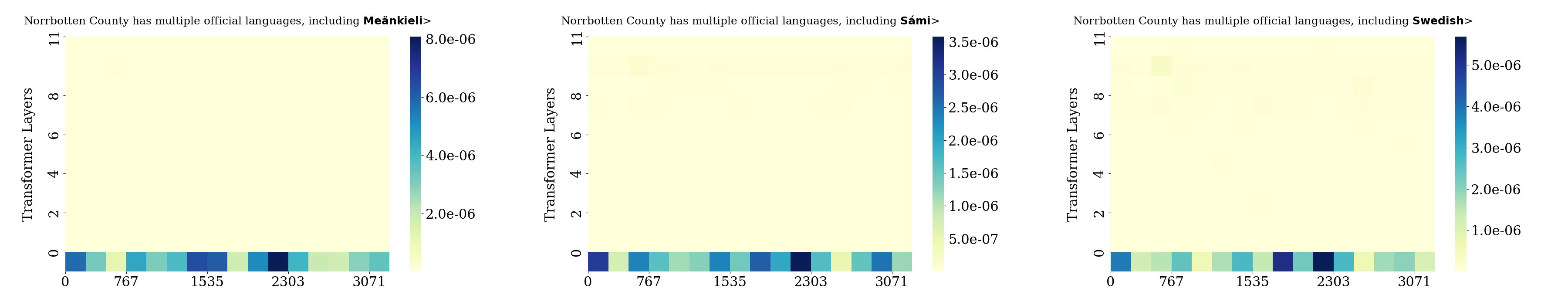}
   \end{center}
   \caption{\label{fig:case}
   Analysis for knowledge overwriting using case 3.
   }
\end{figure*}

\begin{figure*}[h]
   \begin{center}
   \includegraphics[width=1\linewidth]{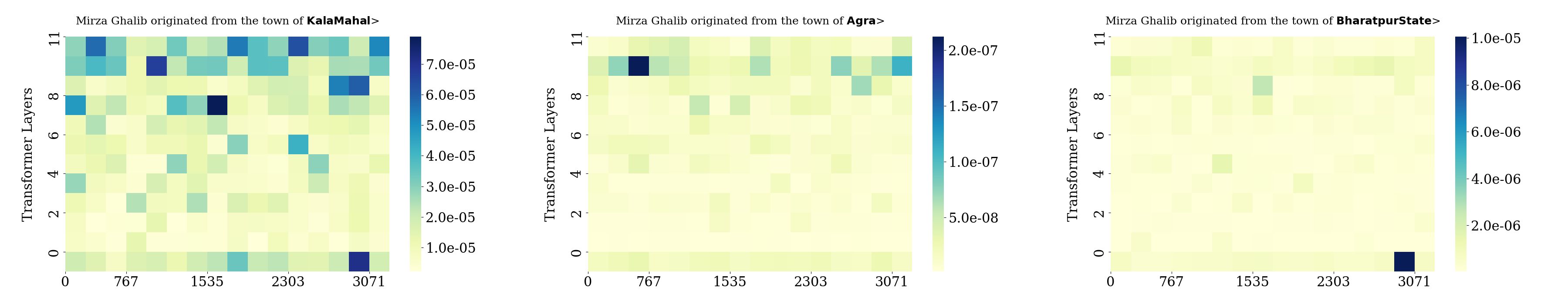}
   \end{center}
   \caption{\label{fig:case}
   Analysis for knowledge overwriting using case 4.
   }
\end{figure*}

\begin{figure*}[h]
   \begin{center}
   \includegraphics[width=1\linewidth]{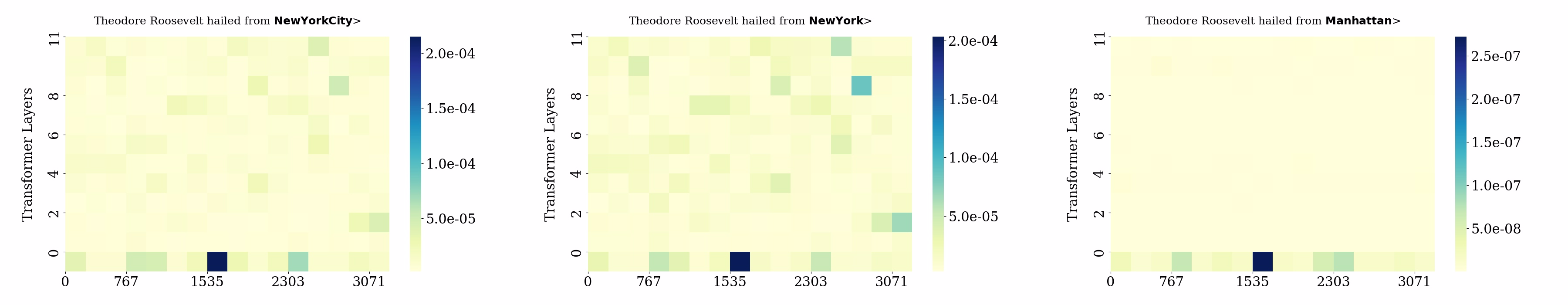}
   \end{center}
   \caption{\label{fig:case}
   Analysis for knowledge overwriting using case 5.
   }
\end{figure*}

\begin{figure*}[h]
   \begin{center}
   \includegraphics[width=1\linewidth]{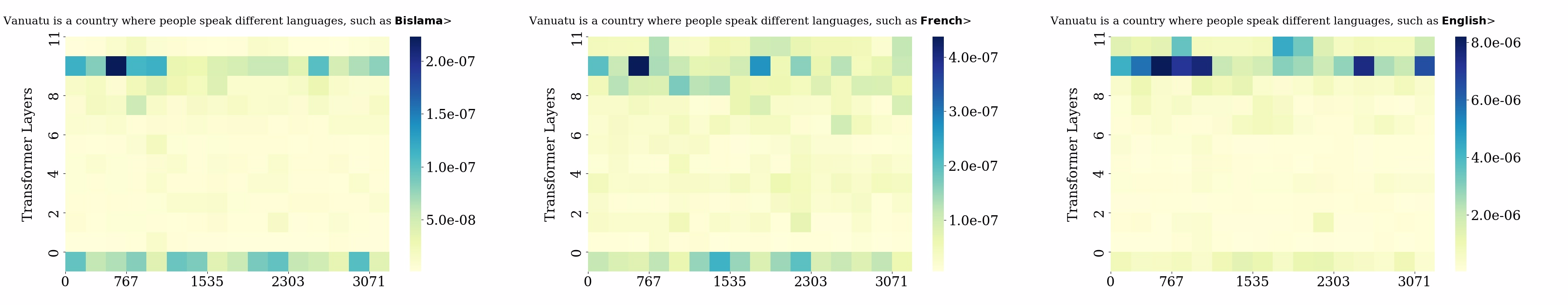}
   \end{center}
   \caption{\label{fig:case}
   Analysis for knowledge overwriting using case 6.
   }
\end{figure*}

\begin{figure*}[h]
   \begin{center}
   \includegraphics[width=1\linewidth]{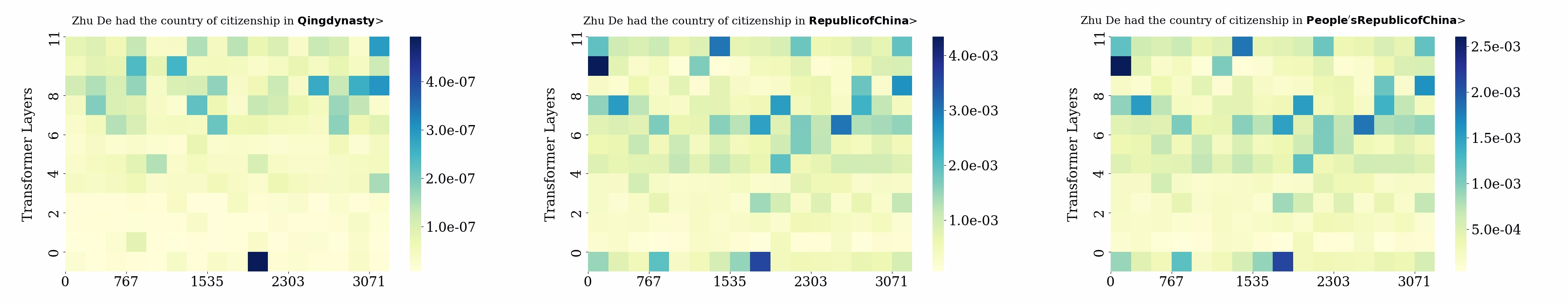}
   \end{center}
   \caption{\label{fig:case}
   Analysis for knowledge overwriting using case 7.
   }
\end{figure*}

\begin{figure*}[h]
   \begin{center}
   \includegraphics[width=1\linewidth]{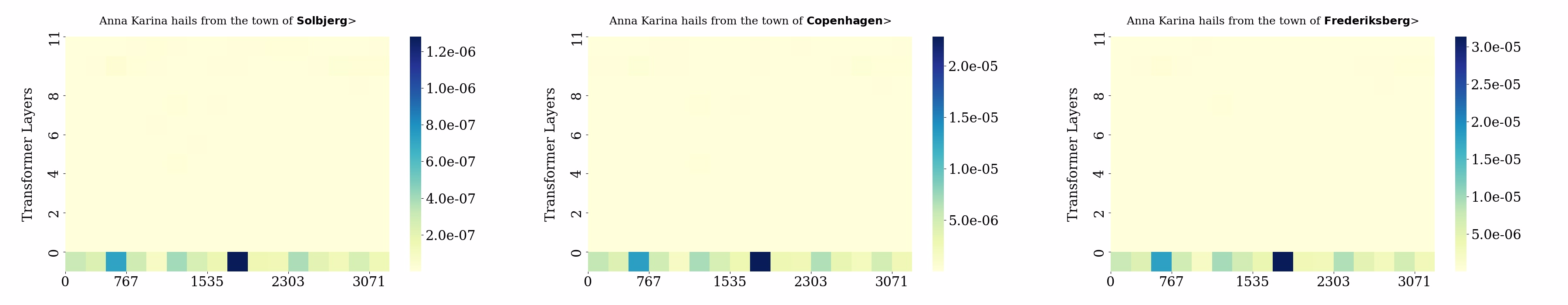}
   \end{center}
   \caption{\label{fig:case}
   Analysis for knowledge overwriting using case 8.
   }
\end{figure*}

\begin{figure*}[h]
   \begin{center}
   \includegraphics[width=1\linewidth]{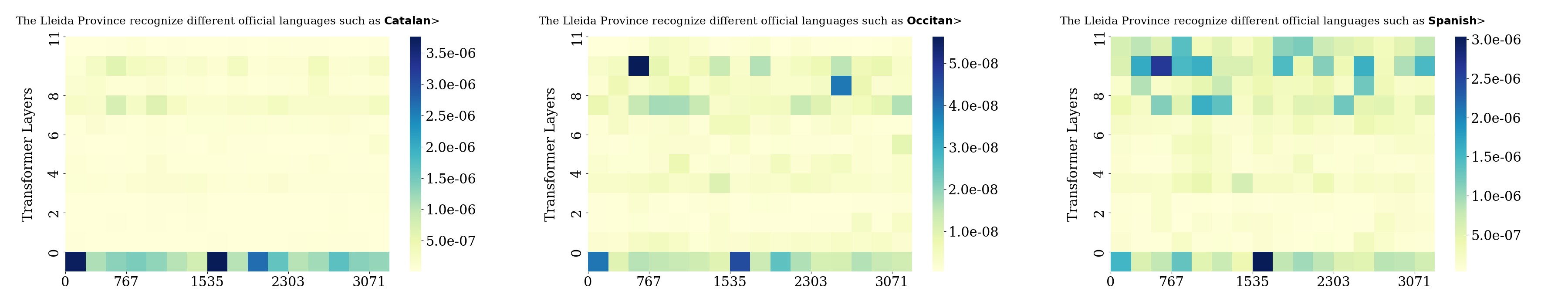}
   \end{center}
   \caption{\label{fig:case}
   Analysis for knowledge overwriting using case 9.
   }
\end{figure*}

\begin{figure*}[h]
   \begin{center}
   \includegraphics[width=1\linewidth]{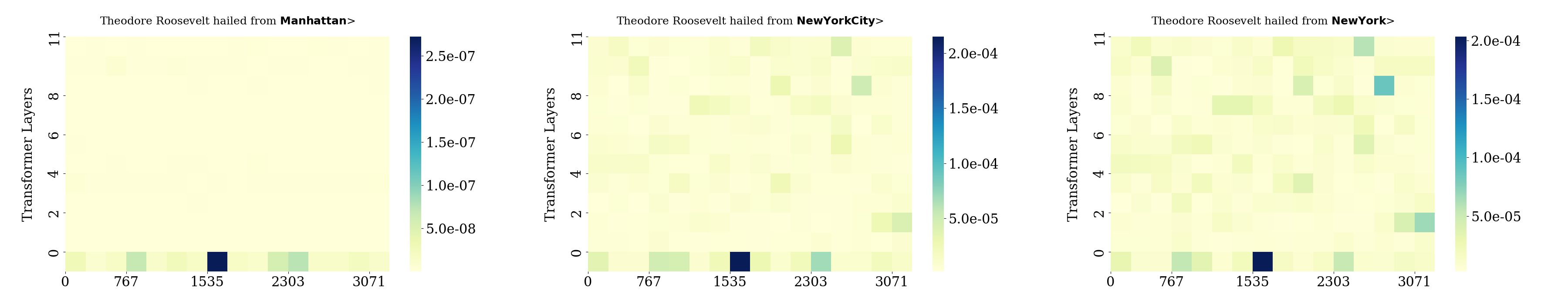}
   \end{center}
   \caption{\label{fig:case}
   Analysis for knowledge overwriting using case 10.
   }
\end{figure*}

\begin{figure*}[h]
   \begin{center}
   \includegraphics[width=1\linewidth]{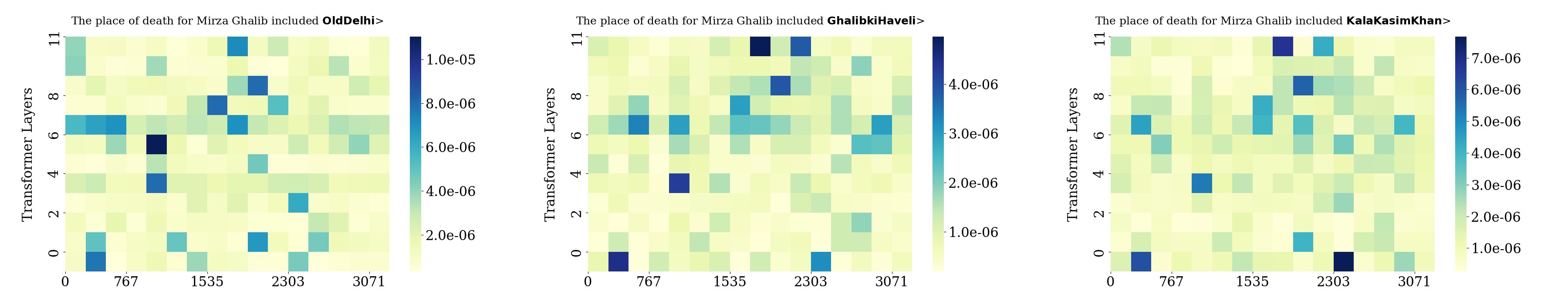}
   \end{center}
   \caption{\label{fig:case}
   Analysis for knowledge overwriting using case 11.
   }
\end{figure*}

\begin{figure*}[h]
   \begin{center}
   \includegraphics[width=1\linewidth]{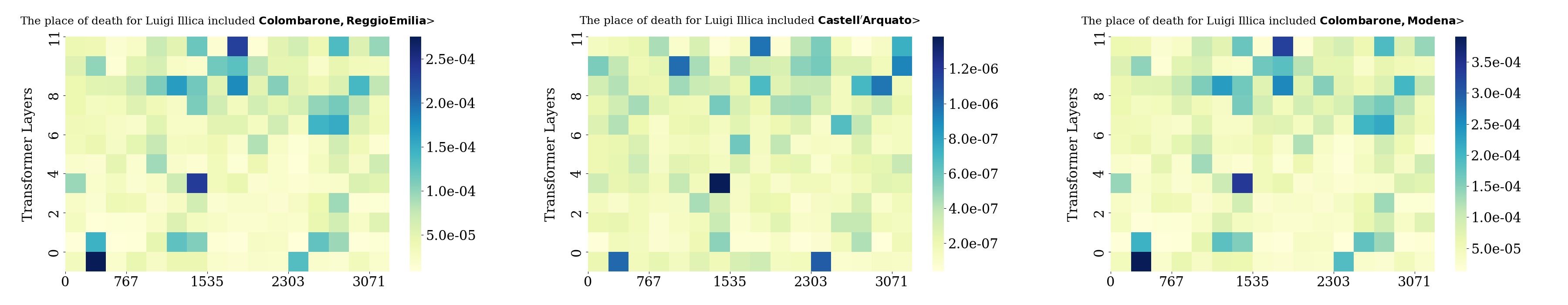}
   \end{center}
   \caption{\label{fig:case}
   Analysis for knowledge overwriting using case 12.
   }
\end{figure*}

\begin{figure*}[h]
   \begin{center}
   \includegraphics[width=1\linewidth]{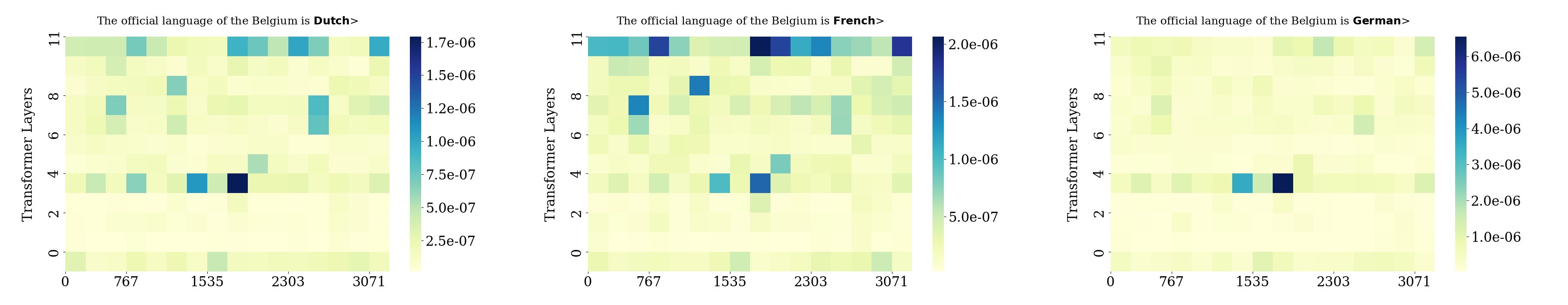}
   \end{center}
   \caption{\label{fig:case}
   Analysis for knowledge overwriting using case 13.
   }
\end{figure*}

\begin{figure*}[h]
   \begin{center}
   \includegraphics[width=1\linewidth]{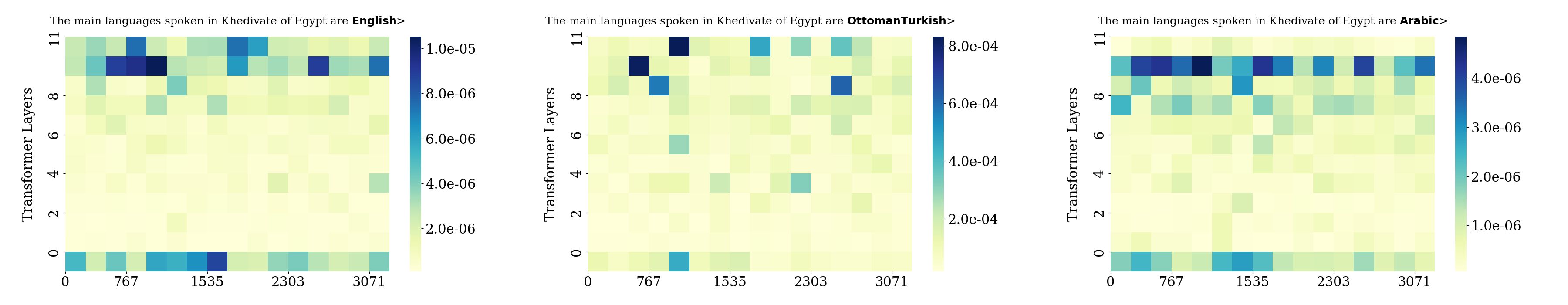}
   \end{center}
   \caption{\label{fig:case}
   Analysis for knowledge overwriting using case 14.
   }
\end{figure*}

\begin{figure*}[h]
   \begin{center}
   \includegraphics[width=1\linewidth]{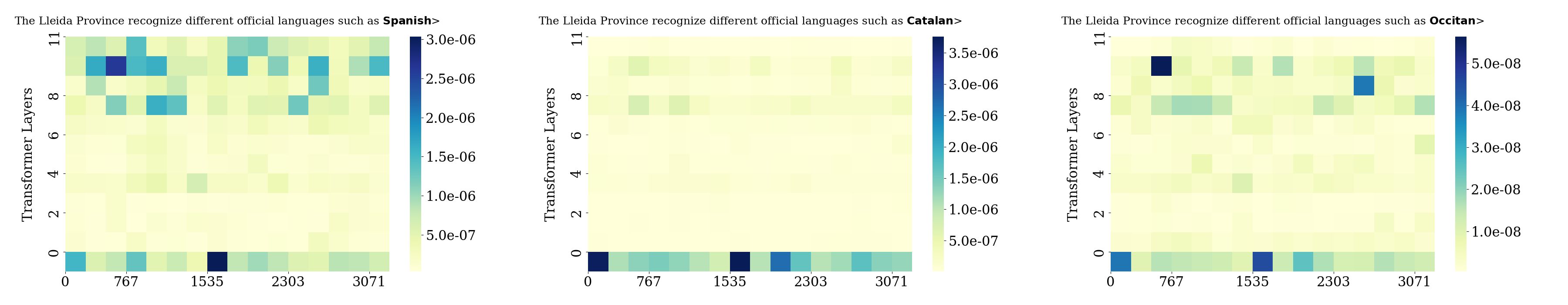}
   \end{center}
   \caption{\label{fig:case}
   Analysis for knowledge overwriting using case 15.
   }
\end{figure*}

\begin{figure*}[h]
   \begin{center}
   \includegraphics[width=1\linewidth]{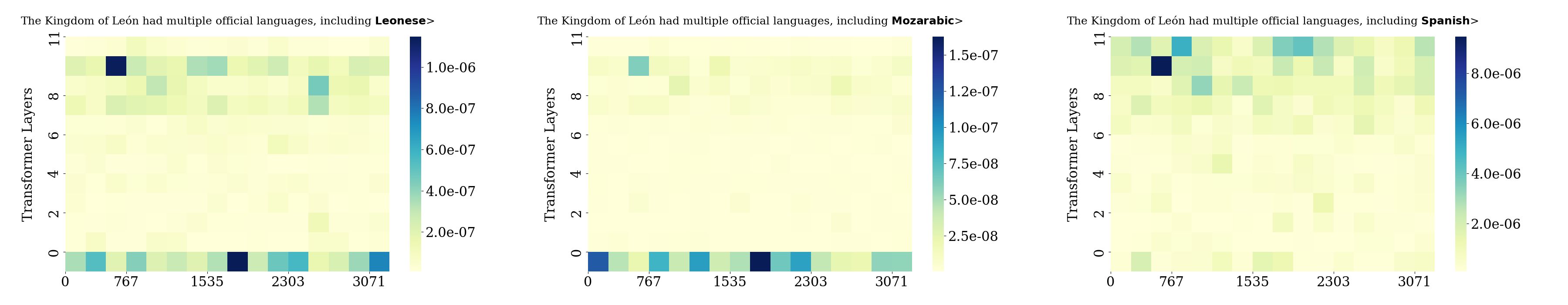}
   \end{center}
   \caption{\label{fig:case}
   Analysis for knowledge overwriting using case 16.
   }
\end{figure*}

\begin{figure*}[h]
   \begin{center}
   \includegraphics[width=1\linewidth]{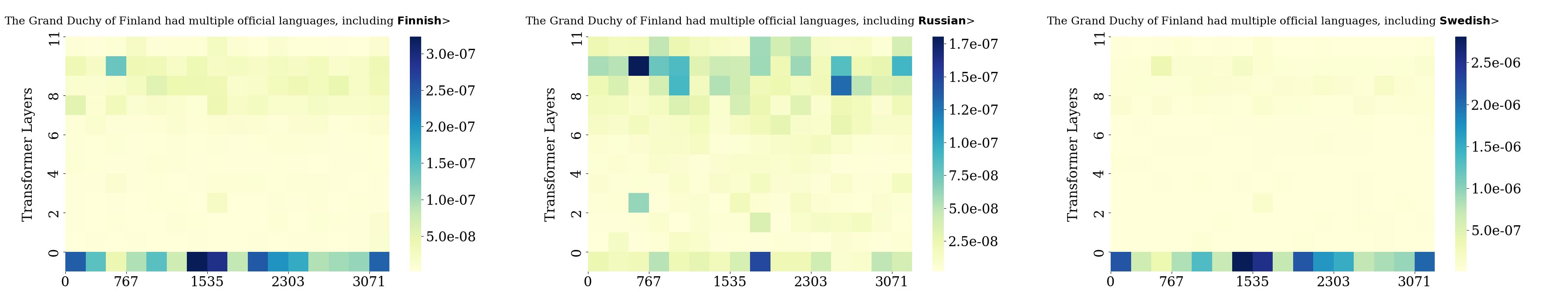}
   \end{center}
   \caption{\label{fig:case}
   Analysis for knowledge overwriting using case 17.
   }
\end{figure*}

\begin{figure*}[h]
   \begin{center}
   \includegraphics[width=1\linewidth]{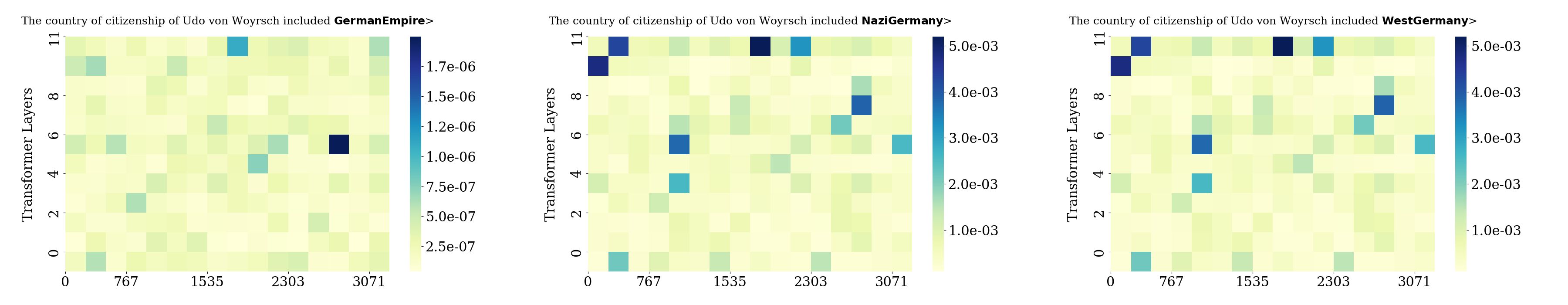}
   \end{center}
   \caption{\label{fig:case}
   Analysis for knowledge overwriting using case 18.
   }
\end{figure*}

\begin{figure*}[h]
   \begin{center}
   \includegraphics[width=1\linewidth]{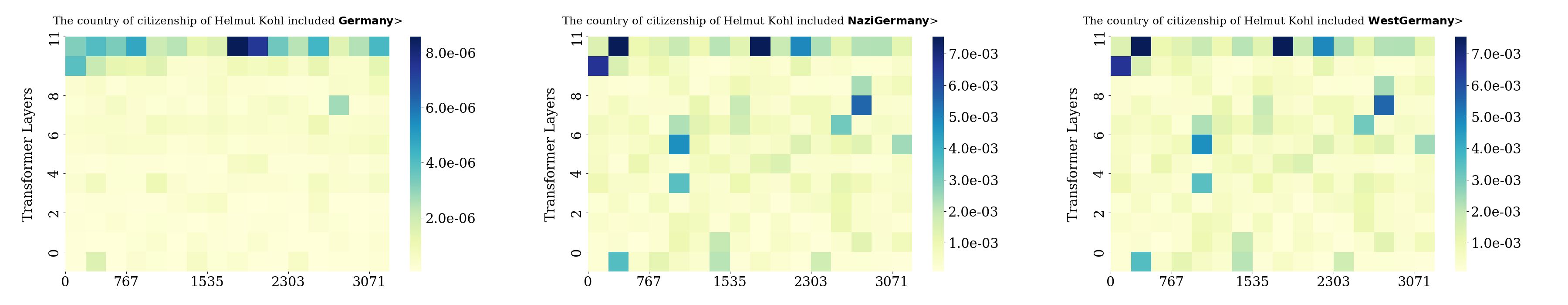}
   \end{center}
   \caption{\label{fig:case}
   Analysis for knowledge overwriting using case 19.
   }
\end{figure*}

\begin{figure*}[h]
   \begin{center}
   \includegraphics[width=1\linewidth]{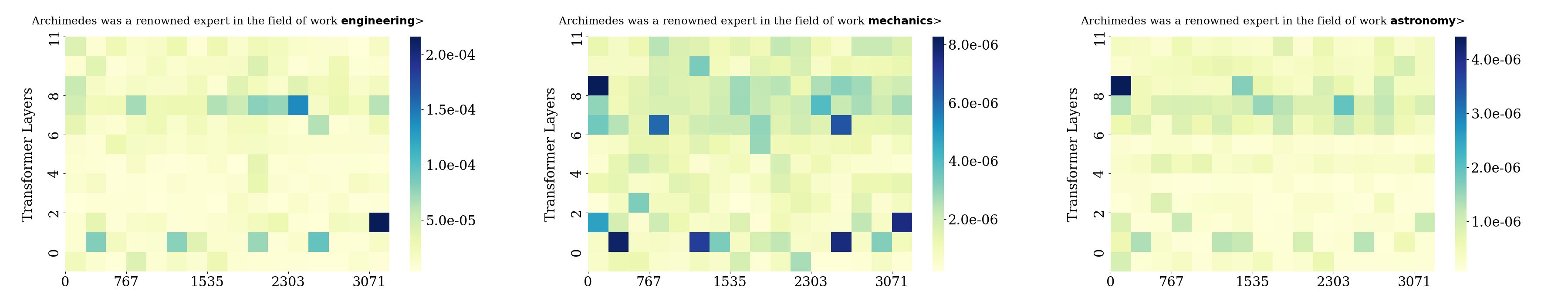}
   \end{center}
   \caption{\label{fig:case}
   Analysis for knowledge overwriting using case 20.
   }
\end{figure*}

\end{document}